\documentclass[letterpaper, 10 pt, conference]{ieeeconf}  % Comment this line out if you need a4paper

\IEEEoverridecommandlockouts
\usepackage{cite}
\usepackage{amsmath,amssymb,amsfonts}
\usepackage{algorithmic}
\usepackage{graphicx}
\usepackage{textcomp}
\usepackage{balance}
\usepackage{xcolor}
\usepackage{xcolor,colortbl}
\usepackage{multirow}
\usepackage{placeins}
\usepackage{caption} 
\def\BibTeX{{\rm B\kern-.05em{\sc i\kern-.025em b}\kern-.08em
    T\kern-.1667em\lower.7ex\hbox{E}\kern-.125emX}}
\begin{document}

\title{Humans Social Relationship Classification during Accompaniment
}

\author{Oscar Castro, Ely Repiso, Anaís Garrell and Alberto Sanfeliu% <-this % stops a space
%\thanks{*Work supported by the Spanish State Research Agency through the Mar\'ia de Maeztu Seal of Excellence to IRI (MDM-2016-0656)}% <-this % stops a space
\thanks{The authors are with the Institut de Rob\`otica i Inform\`atica Industrial (CSIC-UPC). Llorens Artigas 4-6, 08028 Barcelona, Spain.
        {\tt\small
        oscar.castro.arcusa@estudiantat.upc.edu,  \{erepiso,agarrell,sanfeliu\}@iri.upc.edu}}%
\thanks{}%
}

%Work supported by the Spanish State Research Agency through the Mar\'ia de Maeztu Seal of Excellence to IRI (MDM-2016-0656)
% Work supported by the Spanish Ministry of Science and Innovation under project ROCOTRANSP project (PID2019-106702RB-C21 / AEI / 10.13039/501100011033), the EU project TERRINET (H2020-INFRAIA-2017-1-two-stage-730994) and by the Spanish State Research Agency through the Mar\'ia de Maeztu Seal of Excellence to IRI (MDM-2016-0656)

%\author{\IEEEauthorblockN{1\textsuperscript{st} Óscar Castro Arcusa}
%\textit{name of organization (of Aff.)}\\  % organizacion o IRI o upc (yo supongo que tendré que poner la del LAAS, creo...
%ocastro@iri.upc.edu} %tambien puedes poner la de la upc si quieres.
%\and
%\IEEEauthorblockN{2\textsuperscript{nd} Given Name %Surname}
%\and
%\and
%\and

%}

\maketitle

\begin{abstract}
This paper presents the design of deep learning architectures which allow to classify the social relationship existing between two people who are walking in a side-by-side formation into four possible categories --colleagues, couple, family or friendship. The models are developed using Neural Networks or Recurrent Neural Networks to achieve the classification and are trained and evaluated using a database of readings obtained from humans performing an accompaniment process in an urban environment. The best achieved model accomplishes  a relatively good accuracy in the classification problem and its results enhance partially the outcomes from a previous study~\cite{yucel2019identification}. Furthermore, the model proposed shows its future potential to improve its efficiency and to be implemented in a real robot.
\end{abstract}

% keywords de francesco: dyads, interaction, recognition, pedestrian groups, social relation

\section{Introduction}
% TODO, buscar alguna imagen de gente caminado con tibi para que se vea el objetivo principal.
% Las tres preguntas que toda intro ha de contestar:
%%%%%%%%%%%%(1) cual es el problema? what is the problem?
%%%(2) Why is this work important??
%%%%%%%%%%%%%%%%%(3) porque es dificil? Why is it difficult?

% porque es importante:
% porque es importante + porque es dificil? (pero faltaria desarrollar algo más el pq es dificil). Creo que lo incluye bien el porque es dificil, aunque no se diga explicitamente.

The world of robotics is experiencing an unprecedented growth towards artificial intelligence or big data. Also, the field of Human Robot Interaction (HRI) is no exception in this regard, more and more researchers include these theories to allow the development of robots capable of executing more natural, safe, social and comfortable tasks for  humans who interact with them to perform everyday tasks~\cite{lemaignan2017artificial,doering2021data, villamizar2016interactive}.  % si se os ocurren otros, up to you. Yo son estos los más punteros que se ahora mismo que existen.  
%Furthermore, there is tendency to allow robots to share the same space, interact and collaborate with humans to do everyday tasks. 

Furthermore, this interaction between robots and humans is presented as one of the great challenges for robotics that must be faced, and even more so if we want non-trained volunteers to feel comfortable interacting with these robots. For this reason, we must equip robots with more human and social skills, like the accompaniment task which is present in many situations, such as shopping~\cite {gross2009toomas}, traveling to other places~\cite {hirose2015personal}, or visiting museums~\cite{faber2009humanoid}. Where the robot should be adapted to allow a more human, predicable and comfortable accompaniment behaviors. To do so, the classification of these human social relationships is crucial to allow the robot to behave in consequence.

%robots need to classify these different interactions and behave depending on the social relation that shows the their human partners. %In the same way that humans interact differently depending on the person they are interacting with, each human will behave differently depending on the robot they are interacting with, and the type and time of interaction.

%No, esto sería más de cara a si estubiera implementado en un robot real: Esta tarea supone un gran desafio para los robots ya que implica ser capaces de aprender del comportamiento huma diferentes tareas como: predecir el comportamiento de su compañero, adaptarse a su compañero, seguir las normas sociales de las peronas, realizar comportamientos predecibles y naturales para las personas.

%%%%%%%%%%%%(1) cual es el problema? what is the problem?

In this research we are interested in the development of social robots capable of accompanying pedestrians. 
To ensure this purpose, the robot which is accompanying a person should adapt its behavior in the best possible way,  and it is mandatory to  be able to analyze, to learn and to execute the variables that the accompanying processes imply to give an adequate response to each person and situation, in order to obtain a comfortable and safe interaction for both parties. 
Thanks to the correct identification using the human social relationship classification during the accompaniment task, the robot should be able to adapt its movements and to generate a greater sense of security and comfort for people and, therefore, improve their experience during the accompaniment process.

%En primer lugar, es necesaria la realización de un programa capaz de clasificar la relación entre dos personas a partir de las características observables durante un proceso de acompañamiento a lo largo del tiempo para, posteriormente, poder usar ese aprendizaje en los robots. Dicho aprendizaje les permitirá clasificar el comportamiento humano durante los primeros instantes de la interacción y poder así obtener un mejor y más personalizado acompañamiento de personas.

\begin{figure}[t]
\begin{center}
\includegraphics[width=.4\textwidth]{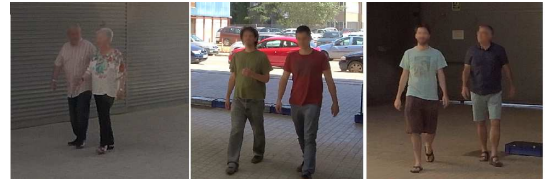} %pareja_amigos_colegas
 %Intro_paper_solo_pers2  (0.25)
\end{center} % Imagen_Intro_paper ==> sin blurred
 %VFORM-AGROUP  and  SIDE-AGROUP
\caption{ {\small The main objective of this research is to  obtain the classification of the social relationship of humans being accompanied by each other. To make  a robot capable of classifying if they are couple, friends or colleagues.% to allow, in a future, that the robot should be able to adapt its behavior to the social relation showed by its human partner.
} }
%\vspace{-1mm}
\label{fig:intro_paper}
\vspace{-4mm}
\end{figure}

In this paper, we present new methods to classify the social relationship of humans being accompanied by other humans, to allow a robot to learn these types of social behaviors so that in the future, it would be capable of accompanying  people in a social and more comfortable manner. Fig.~\ref{fig:intro_paper} shows a set examples of different relationships between  humans navigating in a side-by-side formation.

%\TODO{Falta parrafo que comente que se describe en cada sección brebemente. Eso mejor incluirlo cuando esten claras las secciones.}

In the remainder of the paper, the related work is presented in Sec.~\ref{sec:related_work}. Sec.~\ref{sec:method} describes the used database and exposes the implemented architectures of Neural Networks (NNs) and Recurrent Neural Networks (RNNs). Sec.~\ref{results} shows the obtained results.  Sec.~\ref{Discussion} includes the discussion of the paper. Finally, conclusions are given in Sec.~\ref{sec:conclusions}.

\section{Related Work}
\label{sec:related_work}
The main purpose of the current research is to make robots capable of accompanying a person in a social accepted manner. However, to achieve this goal, robots should understand the relationship established between individuals during the task, in order to adapt their behavior. For this reason, it is necessary, in the first place, to classify the type of affinity that is constituted between two people, so that in the future it could be applied to the robot-human relationship.

\subsection{Accompaniment with social robots}
\label{Accompaniment}

One of the most promising applications in recent years in the field of social robots is accompanying people. In society, there are a wide variety of situations in which a social robot can carry out accompaniment tasks, whether accompanying a group of humans or a single person. In particular, the ability to accompany a single person is an important tool that every urban robot should have, responding to the basic need to accompany people in a natural and safe way. Some of the situations where side-by-side accompaniment is a fundamental pillar arise when guiding~\cite{garrell2012cooperative,yousuf2012development,faber2009humanoid} or accompanying walking people~\cite{morales2014walking,karunarathne2018model}, especially in urban~\cite{repiso2017line} or aerial~\cite{garrell2019teaching} environments.  % este: repiso2020collaborative cambiado por este: repiso2017line. Citas de tesis en paper no molan!

% Human accompaniment is a field that encompasses a wide variety of robotics disciplines, involving a mixture of subjects such as perception, navigation and human-robot interaction (HRI). Despite the heterogeneity of the subjects involved, the task of companionship must be studied holistically, which makes it a particularly complicated task~\cite{repiso2017line}.

% Specifically, HRI research in the field of robot navigation while accompanying humans is still new compared to other traditional fields of robotics, such as traditional robotic navigation, where robots navigate in a safer and more human-like manner. This field presents a challenging task as the robot has to navigate safely and naturally while accompanying a person. In addition, the robot has to adapt its behaviour to avoid collisions with obstacles and not disturb other people in the environment. However, accompanying algorithms are increasingly evolving towards safer and more realistic behaviour~\cite{repiso2020people}.

HRI research in the field of robot navigation while accompanying humans presents a challenging task as the robot has to navigate safely and naturally while accompanying a person. Currently, there are models that allow a robot to navigate with social awareness and accompany a person side-by-side in crowded urban areas~\cite{repiso2017line,repiso2020people,prassler2002key}, even without knowing the destination of the accompanied person~\cite{karunarathne2018model}. However, not all existing models are able to adjust certain parameters, such as social distance or accompanying speed, according to the specific characteristics of the accompaniment or the particular environment of the accompanied person~\cite{morales2014walking}.

In addition, it has been shown that there are several factors that affect the social interactions between the robot and the accompanied person, e.g. if the accompanied people have experience in pet care or in the world of robots, they decrease their personal space with respect to the robot compared to the space that inexperienced people maintain~\cite{takayama2009influences}.

%In today's society, there are a wide variety of situations in which a social robot could carry out tasks of accompanying a person: acting as a guide in an unfamiliar space~\cite{yousuf2012development,faber2009humanoid}, helping to carry objects~\cite{yamashita2000motion}, helping health workers to carry medicines in spaces that are delicate or vulnerable for other humans, etc.

However, the aforementioned works do not consider the different social relationships that are established between individuals in the accompaniment task. In contrast, we present a classification model of human behavior that allows us to identify the social relationship existing between two people in a couple performing an accompaniment process with the future objective that it can be adapted for a robot, so that when it is used it allows the robot to adapt to each person and, consequently, to have a greater acceptance.

\subsection{Classification of  social relationship between couples}
\label{Classification}

% A robot could accompany a target person and even replace one person in a human couple in tasks where two people are needed, for example to save work for one person or to minimise costs, risks or time. In this way, the accompanying robot is expected to act as closely as possible to how a person would act, i.e. following the same social conventions that the person would follow and adapting to the behaviour displayed by the accompanied person. Therefore, the robot must somehow learn to behave as humanely as possible, and studies on the classification of the social relationship between the members of a couple who accompany each other may be key to this~\cite{yucel2019identification}.

% To facilitate the adaptation of the robot to the accompaniment expected by the person accompanied, it is important to try to relate the variables of the accompaniment processes between two people to the social relationship that these two people have with each other.

% (relative speeds of the people, distance between them, absolute speed of the group, etc.)

In general, there are a wide variety of studies focused on classifying the social relationship between people with the aim of designing machines and robots capable of interacting socially with humans~\cite{wang2018deep,sun2017domain,li2020visual}. Focusing on environments such as urban areas, there are also studies that focus on classifying the social relationship between members of pedestrian groups in public spaces~\cite{yucel2020estimating}, or analysing how the type of relationship between members of the same social group affects the dynamics of the group when moving around~\cite{zanlungo2019intrinsic}. 

For example, authors in~\cite{yucel2019identification} presents a study that focuses on analysing social groups of pedestrians in public spaces with the aim of identifying the type of relationship between group members. In particular, the study uses pedestrian couples as the social group of reference.

Although there are various classifications of human-to-human social relationships~\cite{fiske1992four,clark1979interpersonal,foa2012resource}, the above study~\cite{yucel2019identification} classifies the relationship of pedestrian couples into four categories: colleagues, couple, family or friends~\cite{bugental2000acquisition}. Using two different methods, the study manages to obtain relatively good results in classifying the relationship of the couples and distinguishing between the four categories above. Also, we compare our results with this work.

In contrast to previous studies, this work develops different models that perform the classification of the existing social relationship in couples during accompaniment based on the application of deep learning models. Therefore, the methods implemented allow for the classification of the relationship between the members of a couple while they are in the process of accompaniment into four possible categories:  %depending on whether their social relationship, is that of
colleagues, couple, family or friendship. Specifically, we implement various models that use NNs, both standard and recurrent, to perform the classification. 
%methods based on deep learning implement various models that use neural networks, both standard and recurrent, to perform the classification.

\section{Classification of Humans' Relationship}%methods
\label{sec:method}

For social robots to find their place in today's society in the field of accompanying people, it is of vital importance that they are able to adapt to human behavior when carrying out these accompanying tasks. An indispensable part of these tasks is to observe this behavior and identify what kind of relationship is established with the person being accompanied. %Beforehand, it is necessary to be able to classify the types of relationships that humans have between pairs or groups of two people.

In order that the robot will be able to classifying social relationships between humans, we describe the database that we use in Sec.~\ref{Database_explanation}, and deep learning is used to create several models to obtain this robot's behavior in Sec.\ref{Neural_Network}. %The goal of deep learning is to model complex representations in data sets using computational architectures known as Neural Networks~\cite{bengio2013representation}.

\subsection{Database Description}
\label{Database_explanation}

In order to design the desired classification method it is need it a database to work on. This database allows deep learning models to be trained and the effectiveness of their classifications to be tested. The used database in this work was provided by Dr. Francesco Zanlungo of the Intelligent Robotics and Communication Laboratory of the Advanced Telecommunications Research Institute (ATR) in Kyoto, author of several studies in the field of HRI and social robots~\cite{zanlungo2019intrinsic,yucel2019identification}. 

This database contains the readings of different variables taken from couples performing an accompaniment process, walking through an urban environment, and the labels corresponding to the couples' social relationships. These labels were assigned by several people that observe the couple behavior and recognize their relation. The provided database  presents the examples of couples divided into four categories --colleagues, couple, family or friendship. In total, it consists of $867$ examples of couples. The distribution of examples of pedestrian couples in the four studied categories are: $267$ in Colleagues, $96$ in Couple, $218$ in Family, and $286$ in Friendship.  %, and we used only the data related with couples interacting with each other.
%Details of the process of obtaining and classifying the data can be read in~\cite{yucel2019identification, zanlungo2015spatial}. 

%Each experiment contains the readings of a single pair. Furthermore, in this work, we only used data from couples interacting with each other.

% The database provided presents the examples of couples divided into four folders, each of the four representing one of the categories of the classification (colleagues, couple, family or friendship). In total, the database provided consists of $867$ examples of couples. Each folder contains the files of the experiments, where each file contains the readings of a single pair. Each file presents in the first lines details of the couple (identifiers of each person in the tracking system, whether they are male or female, age and height). Furthermore, for our work we only used data from couples interacting with each other.

\begin{figure}[t]
\begin{center}
\includegraphics[width=.34\textwidth]{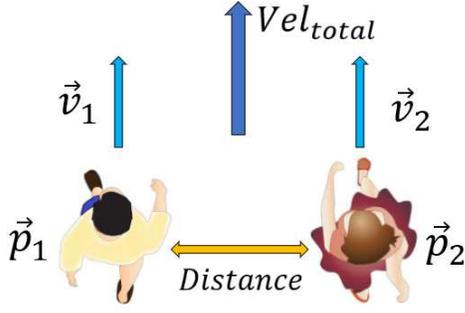} 
\end{center} 
\caption{ {\small Representation of the variables of the accompaniment process obtained from the examples in the database.
} }
\vspace{-3mm}
\label{fig:variables}
\end{figure}

Each example of couple includes $13$ different variables: detection time; $\overrightarrow{p}_c =$($p_X$, $p_Y$, $p_Z$)$_{c}$ where $c \in \{1,2\}$ meaning position of pedestrian $1$ and $2$; $\overrightarrow{V}_c=(V_X,V_y)$  velocity of pedestrian $1$ and $2$; and ${{V}^T}_c$ total velocity of pedestrian $1$ and $2$. Most of these variables can be seen graphically in the Fig.~\ref{fig:variables}. In addition, to the aforementioned $13$ parameters, three new variables are calculated using these previous parameters due to the importance they prove to have in the processes of social accompaniment and navigation~\cite{zanlungo2019intrinsic,yucel2019identification,yucel2020estimating}. These new three variables are included in Eq.~\ref{eq:dist},~\ref{eq:velRel} and~\ref{eq:velTot}. All these variables are used to characterize the performed type of accompaniment. As the accompaniment processes are dynamic processes given during navigation, the experiments have different measurements as each process has a different duration. %Nota, a mi esta ultima frase: As the accompaniment processes... No me aporta nada. Yo la quitaria.

\begin{itemize}
\item	Distance between pedestrians, calculated as

\begin{equation}
Dist = ||{\overrightarrow{p}_2} - {\overrightarrow{p}_1}||, 
\label{eq:dist}
\end{equation}
where $ {\overrightarrow{p}_2} $ is the position of pedestrian $2$ and $ {\overrightarrow{p}_1} $ is the position of pedestrian $1$.

\item	Relative pedestrian-to-pedestrian velocity, calculated as

\begin{equation}
{Vel_{Relative}} = |{{V_1^T} - {V_2^T}}|, 
\label{eq:velRel}
\end{equation}
where $ {V_1^T} $ is the total velocity of pedestrian $1$ and $ {V_2^T} $ is the total velocity of pedestrian $2$.

\item	Total velocity of the couple, calculated as 

\begin{equation}
{Vel_{Total}} =  \frac{{V_1^T} + {V_2^T}}{2},
\label{eq:velTot}
\end{equation}

% where $ {V_1^T} $ is the total velocity of pedestrian $1$ and $ {V_2^T} $ is the total velocity of pedestrian $2$.

\end{itemize}

In summary, the database used is made up of $867$ examples of pedestrian couples, each of which is defined by $16$ parameters that will allow the developed method to classify each example in one of the defined categories. Further details about the database can be obtained in~\cite{yucel2019identification, zanlungo2015spatial}. %Specifically, %Table~\ref{tabla:Distribución_ejemplos} shows the distribution of the $867$ examples in the database in the four categories studied. 

%\begin{comment}
%\begin{table}[t]
%\begin{center}
%\begin{tabular}{ | c | c | }
%\hline
%\textbf{Category} & \textbf{Nº of examples}\\
%\hline
%\textbf{Colleagues} & 267 \\ \hline
%\textbf{Couple} & 96 \\ \hline
%\textbf{Family} & 218 \\ \hline
%\textbf{Friendship} & 286 \\ \hline
%\hline
%\textbf{Total} & 867 \\ \hline
%\end{tabular}
%\caption{\small{Distribution of the examples of pedestrian couples in the database in the four studied categories.}}
%\label{tabla:Distribución_ejemplos}
%\end{center}
%\end{table}
%\end{comment}

\subsection{Database adaptation for our Deep Learning Architectures}

As couple accompaniment is a dynamic process over time, the readings of variables at a given instant may not accurately represent the reality of the relationship. To do this, we took the readings from each experiment and averaged each variable, converting the readings from one experiment into a single reading, except for the time variable, which is obtained by subtracting the final and initial times.

To train the different NNs designed, the existing database is divided into training set and test set. Thus, $90$ \% of the data is used as a training set to train the model and the remaining $10$ \% is used to test the accuracy and validity of the trained model as a test set. Therefore, if the database consists of $867$ experiments, a total of $780$ examples make up the training set and the remaining $87$ examples are part of the test set.

\subsection{Deep Learning Architecture}
\label{Neural_Network}

Neural networks (NNs) are especially notable for their ability to process information due to their structure. They are built by complex computational units, known as neurons, which are distributed in layers and are interconnected forming a network of elements that work together to solve specific problems~\cite{abiodun2018state}.

The proposed classification methods are based on NNs. The design of the proposed models is based on finding the correct configuration of the so-called hyperparameters of the system. These hyperparameters are those variables of the model that must be configured and adjusted during the design and that allow controlling the training process of the NN. 

% Some of the hyperparameters used in the design and optimization of NN are: number of hidden layers, number of neurons in the hidden layers, mini-batch size, number of epochs, choice of activation function, choice of learning rate or regularization implementation.

Different NN designs are used to test which one offers the best accuracy, both in the training set and, especially, in the test set. In each design implemented, several hyperparameters of the network are varied: the number of hidden layers, the number of neurons in the hidden layers, the learning rate and the number of epochs in the training process. In addition, the L2 regularisation and dropout methods are used to try to solve the overfitting problem.

There are several design conditions that are met for all NNs developed. These conditions are:

\begin{itemize}

\item The input layer consists of $16$ neurons. Each neuron corresponds to one of the input parameters explained in the subsection~\ref{Database_explanation}. 

\item The hidden layers are designed with the \textit{ReLU} activation function.

\item The output layer is designed with the \textit{Softmax} activation function.

\item The Mini-Batch Gradient Descent is used as gradient descent algorithm.

\item The croos-entropy function is used as the loss function.

\end{itemize}

\begin{figure}[t]
\begin{center}
\includegraphics[width=.35\textwidth]{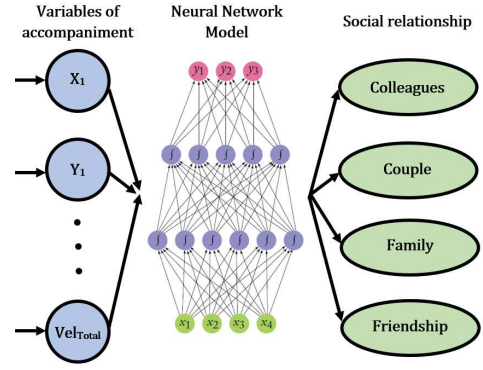} 
\end{center} 
\caption{ {\small General outline of the developed classification method based on deep learning architectures. On the left side are the variables of the accompaniment process that will be processed by the NN example~\cite{goldberg2017neural}. At the output of the network, the predicted social relationship will be obtained.
} }
\label{fig:overview}
%\vspace{-0,5mm}
\end{figure}

A general outline of the developed classification method can be seen in Fig.~\ref{fig:overview}. On the one hand, the variables extracted from the examples in the database feed the deep learning models designed. On the other hand, the classification of the example into one of the defined social relationship types is obtained in the NN output.

\subsection{Recurrent Neural Network}
\label{Recurrent_Neural_Network}

A recurrent neural network (RNN) is a type of NN capable of working with temporal sequences of data. The ability to have memory makes RNNs a suitable tool for machine learning tasks involving sequential data. By being able to use relevant information from past input data in the training process, they can make more accurate predictions~\cite{li2018independently}.

Once the standard NN models have been implemented, several RNNs are designed to test their efficiency. Having readings of the database samples over a certain period of time, it is decided to implement RNNs due to their ability to work with temporal sequences of data and extract possible dependencies between them.

Specifically, Long Short-Term Memory (LSTM) networks are an extension of RNNs that use special hidden units, called memory cells, whose goal is to increase the memory of the network so that it is able to remember important information over time~\cite{yu2019review}. In this way, LSTMs can better capture temporal dependencies of long-term input data, due to the ability of LSTM units to register possible temporal behavioral patterns. In addition, they have the ability to avoid the phenomenon of vanishing or exploding gradients~\cite{torres2018deep}.

The RNNs developed in this work follow the same design conditions as NNs, as explained in subsection~\ref{Neural_Network}. In addition, the first hidden layer of the RNNs is a LSTM layer, which allows the model to capture the temporal dependencies of the input data in the long term.

\section{Results}
\label{results}

\begin{table}[t]
\begin{center}
\begin{tabular}{ | c | c | c | c | c | c | c | c | c | }
\hline
\parbox{0,25cm}{\textbf{}} & \parbox{0,83cm}{\textbf{Nº of hidden layers}} & \parbox{1,4cm}{\textbf{Nº of \\ neurons in \\ the hidden layers}} & \parbox{0,65cm}{\textbf{Nº of epochs}} & \parbox{0,96cm}{\textbf{Learning \\ rate}} & \parbox{0,35cm}{\textbf{L2 \\ Reg.}} & \parbox{0,85cm}{\textbf{Dropout}}\\
\hline
\textbf{RN2-1} & 2 & 25-12 & 1500 & 0,00015 & No & No \\ \hline
\textbf{RN2-2} & 2 & 1500-600 & 2500 & 0,00011 & No & No \\ \hline
\textbf{RN2-3} & 2 & 1500-600 & 2500 & 0,00011 & Yes & No \\ \hline
\textbf{RN2-4} & 4 & \parbox{1,4cm}{1800-2500-\\ 1600-600} & 2500 & 0,00011 & Yes & No \\ \hline
\textbf{RN2-5} & 2 & 1500-600 & 2500 & 0,00011 & Yes & Yes \\ \hline
\end{tabular}
\caption{\small{Features of standard NN models implemented.}}
\vspace{-6mm}
\label{tabla:ResumenTF2-RN-Parámetros}
\end{center}
\end{table}

\begin{table}[t]
\begin{center}
\begin{tabular}{ | c | c | c | c | c | c | c | c | c | }
\hline
\textbf{} & {\textbf{Training set accuracy}} & {\textbf{Test set accuracy}}\\
\hline
\textbf{RN2-1} & 44,49\% & 26,44\% \\ \hline
\textbf{RN2-2} & 94,10\% & 33,33\% \\ \hline
\textbf{RN2-3} & 99,10\% & 40,23\% \\ \hline
\textbf{RN2-4} & 95,90\% & 36,78\% \\ \hline
\textbf{RN2-5} & 63,08\% & 33,33\% \\ \hline
\end{tabular}
\caption{\small{Accuracy of standard NN models implemented.}}
\vspace{-4mm}
\label{tabla:ResumenTF2-RN-Resultados}
\end{center}
\end{table}

% Having studied the two groups of classification methods to be used in the process of classifying the social relationship in pairs, it is necessary to apply them to the database to check the accuracy offered by each of them in order to select the most effective classification method.
In this section, we present the results obtained form the different neural networks implemented. First, the NNs (of Sec.~\ref{Neural_Network}) and the RNNs (Sec.~\ref{Recurrent_Neural_Network}) models are used to classify the people behavior in four categories in Sec.~\ref{Results_of_NNs_models} and second, the same methods are used to classify the behaviors in only two categories, because the categories of couple, family, and friendship can be merged in one "intimate", due to the fact that they have a similar degree of intimacy, very different than the one of the colleagues category. %XXX(TODO, find our motivation.)

\subsection{Results of NNs models}
\label{Results_of_NNs_models}

%In this section, we present the results obtained using different models of deep learning. To begin with, the classification results obtained by the NNs explained in the subsection~\ref{Neural_Network} are collected.

The characteristics of the NNs implemented are detailed in Table~\ref{tabla:ResumenTF2-RN-Parámetros}. By running the networks defined in Table~\ref{tabla:ResumenTF2-RN-Parámetros}, the results are obtained as shown in Table~\ref{tabla:ResumenTF2-RN-Resultados}. To try to avoid the phenomenon of overfitting, in some models L2 regularisation and dropout methods ($15$ \% in the hidden layers) are implemented. We only show the confusion matrix of the best model selected in table~\ref{tabla:RN2-3-2}. The other confusion matrices show worst results in general or show that the neural net can differentiate well one or two types of the four mentioned, then these neural nets can not be used alone to classify correctly all the people's behaviors. %(\ref{tabla:RN2-1-2},~\ref{tabla:RN2-2-2},~\ref{tabla:RN2-3-2},~\ref{tabla:RN2-4-2},~\ref{tabla:RN2-5-2}).

\begin{table}[t]
\begin{center}
\begin{tabular}{ | c | c | c | c | c | c | }
\cline{3-6}
\multicolumn{2}{c|}{} &\multicolumn{4}{c|}{Predicted value} \\
\cline{3-6}
\multicolumn{2}{c|}{} & \textbf{Colleagues} & \textbf{Couple} & \textbf{Family} & \textbf{Friendship}\\
\hline
\multirow{4}{0,5cm}{Real value} & \textbf{Colleagues} & \cellcolor{orange!34}\textbf{37,50} & \cellcolor{orange!5}4,17 & \cellcolor{orange!23}29,17 & \cellcolor{orange!23}29,17 \\ \cline{2-6}
& \textbf{Couple} & \cellcolor{orange!23}\textbf{28,57} & \cellcolor{orange!23}21,43 & \cellcolor{orange!23}\textbf{28,57} & \cellcolor{orange!23}21,43 \\ \cline{2-6}
& \textbf{Family} & \cellcolor{orange!13}19,05 & \cellcolor{white!0}0 & \cellcolor{orange!44}\textbf{42,86} & \cellcolor{orange!34}38,10 \\ \cline{2-6}
& \textbf{Friendship} & \cellcolor{orange!34}32,14 & \cellcolor{orange!13}10,71 & \cellcolor{orange!5}7,14 & \cellcolor{orange!51}\textbf{50,00} \\ \cline{1-6}
\end{tabular}
\caption{\small{Confusion matrix of the RN2-3 model (in \%).}}
\vspace{-6mm}
\label{tabla:RN2-3-2}
\end{center}
\end{table}

Table~\ref{tabla:RN2-3-2} shows that the examples in the family and friendship categories are correctly classified with accuracies of $42,86$ \% and $50,00$ \%, respectively, when using the RN2-3 model. Also, it may classify part of the other examples of the two other categories, couples and colleagues.

\subsection{Results of RNNs models}

The characteristics of the RNNs implemented are given in Table~\ref{tabla:ResumenTF2-RNR-Parámetros}. The results obtained by executing the networks defined in Table~\ref{tabla:ResumenTF2-RNR-Parámetros} are given in Table~\ref{tabla:ResumenTF2-RNR-Resultados}. In order to try to avoid the phenomenon of overfitting, the L2 regularisation method is implemented in all models. The confusion of the best obtained method is shown in table~\ref{tabla:RNR2-1-2}.

%matrices of the different methods designed are collected in different tables (\ref{tabla:RNR2-1-2},~\ref{tabla:RNR2-4-2},~\ref{tabla:RNR2-5-2}).

Table~\ref{tabla:RNR2-1-2} highlights the accuracies of $82,61$ \% and $30,30$ \% obtained by correctly classifying the examples in the categories of colleagues and friendship, respectively, when using the RNR2-1 model. Also, we can see here that this method tries to classify in the friendship category the categories of family and couple due to the similarity of these three categories. The fact that supports that after, we try to test our classifiers in a new database that join these three types of relation in one.

\subsection{Results analysis}
At first glance, when analysing the results presented in Tables~\ref{tabla:ResumenTF2-RN-Resultados} and~\ref{tabla:ResumenTF2-RNR-Resultados}, it is observed that the best accuracies in the training set are obtained in the models based on standard NNs, as shown in Table~\ref{tabla:ResumenTF2-RN-Resultados}. On the contrary, the best accuracies in the test set are obtained in the models based on RNNs, as shown in Table~\ref{tabla:ResumenTF2-RNR-Resultados}. Specifically, the RN2-3 model, which has an accuracy of $99,10$ \% in the training set and $40,23$ \% in the test set, and the RNR2-3 model, which has an accuracy of $66,52$ \% in the training set and $42,31$ \% in the test set, stand out.

Although the accuracies of the methods in the training set and test set are basic evaluation metrics, an analysis of the confusion matrices obtained is necessary to understand how the classification of the examples into the four categories studied is being performed.

\begin{table}[t]
\begin{center}
\begin{tabular}{| c | c | c | c | c | c | c | c | c |}
\hline
\parbox{0,25cm}{\textbf{}} & \parbox{0,83cm}{\textbf{Nº of hidden layers}} & \parbox{1,4cm}{\textbf{Nº of \\ neurons in \\ the hidden layers}} & \parbox{0,65cm}{\textbf{Nº of epochs}} & \parbox{0,96cm}{\textbf{Learning \\ rate}} & \parbox{0,35cm}{\textbf{L2 \\ Reg.}} & \parbox{0,85cm}{\textbf{Dropout}}\\
\hline
\textbf{RNR2-1} & 2 & 25-12 & 1500 & 0,00015 & Yes & No \\ \hline
%\textbf{RNR2-2} & 2 & 100-50 & 1000 & 0,00011 & Yes & No \\ \hline
%\textbf{RNR2-3} & 2 & 1500-600 & 100 & 0,00011 & Yes & No \\ \hline
\textbf{RNR2-2} & 4 & \parbox{1,4cm}{2500-1800-\\ 1200-600} & 10 & 0,00011 & Yes & No \\ \hline
\textbf{RNR2-3} & 2 & 1500-600 & 25 & 0,00011 & Yes & No \\ \hline
\end{tabular}
\caption{\small{Features of RNN models implemented.}}
\vspace{-2mm}
\label{tabla:ResumenTF2-RNR-Parámetros}
\vspace{-3mm}
\end{center}
\end{table}

\begin{table}[t]
\begin{center}
\begin{tabular}{ | c | c | c | c | c | c | c | c | c | }
\hline
\textbf{} & {\textbf{Training set accuracy}} & {\textbf{Test set accuracy}}\\
\hline
\textbf{RNR2-1} & 45,17\% & 41,03\% \\ \hline
%\textbf{RNR2-2} & 63,64\% & 34,62\% \\ \hline
%\textbf{RNR2-3} & 67,97\% & 38,46\% \\ \hline
\textbf{RNR2-2} & 62,05\% & 30,77\% \\ \hline
\textbf{RNR2-3} & 66,52\% & 42,31\% \\ \hline
\end{tabular}
\caption{\small{Accuracy of RNN models implemented.}}
\vspace{-6mm}
\label{tabla:ResumenTF2-RNR-Resultados}
\end{center}
\end{table}

\begin{table}[t]
\begin{center}
\begin{tabular}{ | c | c | c | c | c | c | }
\cline{3-6}
\multicolumn{2}{c|}{} &\multicolumn{4}{c|}{Predicted value} \\
\cline{3-6}
\multicolumn{2}{c|}{} & \textbf{Colleagues} & \textbf{Couple} & \textbf{Family} & \textbf{Friendship}\\
\hline
\multirow{4}{0,5cm}{Real value} & \textbf{Colleagues} & \cellcolor{orange!70}\textbf{82,61} & \cellcolor{white!0}0 & \cellcolor{orange!5}8,70 & \cellcolor{orange!5}8,70 \\ \cline{2-6}
& \textbf{Couple} & \cellcolor{orange!34}33,33 & \cellcolor{white!0}0 & \cellcolor{white!0}0 & \cellcolor{orange!57}\textbf{66,67} \\ \cline{2-6}
& \textbf{Family} & \cellcolor{orange!44}\textbf{43,75} & \cellcolor{white!0}0 & \cellcolor{orange!13}18,75 & \cellcolor{orange!34}37,50 \\ \cline{2-6}
& \textbf{Friendship} & \cellcolor{orange!57}\textbf{60,61} & \cellcolor{white!0}0 & \cellcolor{orange!5}9,09 & \cellcolor{orange!34}30,30 \\ \cline{1-6}
\end{tabular}
\caption{\small{Confusion matrix of the RNR2-1 model (in \%).}}
\vspace{-4mm}
\label{tabla:RNR2-1-2}
\end{center}
\end{table}

When analysing the confusion matrices of the models based on standard NNs that show better accuracy, different phenomena are observed. In Table~\ref{tabla:RN2-3-2}, corresponding to the model with the highest accuracy in both, the training set and the test set, it is observed that the RN2-3 model recognises the correct categories more accurately than the others, even reaching $50,00$ \% accuracy in the friendship category, and with the exception of the examples in the couples category, which are assigned in a similar proportion among the four categories. This same pattern is repeated in the RN2-4 model%Table~\ref{tabla:RN2-4-2}, corresponding to 
, although the accuracy of the classification of the friendship and couple categories worsens but improves in the family category. The only model that shows high accuracy in the couple category is RN2-1%(see Table~\ref{tabla:RN2-1-2})
, although it shows relatively low efficiency in classifying examples in the other categories. As for the RN2-2 and RN2-5 models, %(see Tables~\ref{tabla:RN2-2-2} and~\ref{tabla:RN2-5-2}, respectively), it is observed that, 
in general, they are not able to classify the examples in the correct categories with a higher accuracy than the others and, moreover, the effectiveness of the classification is in all cases surpassed by that of the RN2-3 model. Thus, the RN2-3 model is postulated as the model based on a standard NN capable of performing a better classification of human relationships in the four categories studied.

On the other hand, when analysing the confusion matrices generated from the models based on RNNs, different phenomena are observed. As can be seen in Table~\ref{tabla:RNR2-1-2}, corresponding to the RNR2-1 model, it is observed that the presented model has a preference when performing the classification to assign the examples in the categories of colleagues or friendship, resulting in a very high accuracy for the category of colleagues ($82,61$ \%), but null for the category of couple and very low for the category of family, because due to its similarity the method classifies all in the category of friendship. The latter is a pattern that is repeated for all the RNN models, as all the models designed have zero accuracy in correctly classifying the examples in the database into the category of couple. Taking this into consideration, the rest of the models present a similar situation by having more facilities to classify the data in certain categories than in others. For example, while the RNR2-3 model is highly accurate in correctly classifying the examples in the colleague category (but with less \% than the selected model), %(see Table~\ref{tabla:RNR2-5-2}), 
the RNR2-2 model is highly accurate in classifying the examples in the family category %(see Table~\ref{tabla:RNR2-4-2}). 
In addition, models based on RNNs have, in general, considerably good accuracy in classifying examples of the colleagues category compared to models based on standard NNs. After all, the RNR2-3 model seems to emerge as the model based on a RNN that can perform a better classification of human relationships by showing, in general, a relatively higher efficiency than the other models in classifying the examples into the four correct categories. However, it has not been possible to produce a model capable of obtaining the best accuracy in at least three of the relationship categories, as the RN2-3 model does.

Therefore, it is considered that the model that can best perform the classification of the social relationship in human couples is the RN2-3 model, as it presents the best accuracy of all the models implemented in the training set, as well as the best accuracies in the test set, and the best efficiencies as a whole when classifying the data correctly (see Table~\ref{tabla:RN2-3-2}).

\section{Conclusions extracted from the implementation of all methods}

The differences between the accuracies obtained in the designed models are due to the different configurations of hyperparameters that have been implemented in them. Depending on these configurations, the networks present higher or lower accuracies in the training set and test set. Even so, when analysing the confusion matrices of the models designed, it is possible to draw a series of conclusions. 

Models with a higher number of neurons in their hidden layers or a higher number of epochs are able to train their network parameters better and, consequently, obtain high accuracies in the training set. However, the accuracy slightly decreases with increasing the number of hidden layers from two to four, as can be seen when comparing the results of the two-hidden-layer RN2-3 model with the results of the four-hidden-layer RN2-4 model (see Table~\ref{tabla:ResumenTF2-RN-Resultados}). This phenomenon may occur as a result of the increased learning difficulty during training due to the increased number of layers. Therefore, an unnecessary increase in the number of layers in the model design should be avoided, because such an increase does not necessarily lead to an increase in model accuracy and, on the contrary, may lead to a reduction in model efficiency.

Another factor to take into account is the training time of the models. When the number of hidden layers, the number of neurons in the hidden layers, or the number of epochs is increased, the computational time and the computational cost of the model training process increases. In the specific case of this work, it has been decided to give priority to the accuracy of the models in the classification, but in the case of looking for a model with a shorter training period, the correct choice of these three hyperparameters must be taken into account in order to achieve this objective.

Moreover, the L2 regularisation method manages to slightly decrease the overfitting phenomenon and increase the accuracy of the model, as can be seen by comparing the results of the RN2-2 and RN2-3 models (see Table~\ref{tabla:ResumenTF2-RN-Resultados}). On the other hand, the dropout method not only fails to increase the accuracy of the model in the test set, but also considerably decreases the accuracy in the training set, as can be seen by comparing the results of models RN2-3 and RN2-5 (see Table~\ref{tabla:ResumenTF2-RN-Resultados}).

Finally, the choice of the learning rate is approached as a process of choosing different values and testing the effectiveness of the models for each of them until the value that offers the best possible results is found.

Consequently, the design of NNs is proposed as an iterative trial-and-error process with the aim of finding the configuration of hyperparameters that offers the best efficiency of the model and, therefore, the best possible accuracy in the database classification process.

Therefore, it can be stated that NNs-based methods of deep learning are positioned as a relatively effective tool for classifying the relationship between people in a couple. However, in order for these methods to realise their full potential, it is necessary to considerably increase the database with more examples of couples, as it is with large datasets that NNs work best. In addition, the process of setting up hyperparameters and testing new NN models should also be continued in order to see how effective they are.

Finally, it can be concluded that the differentiation between the four categories is a complicated process due to the similarity that data from examples of different categories can present, even for a human being (see Fig.~\ref{fig:intro_paper}). In the absence of a behavioral manual for the different social relationships, finding a model capable of identifying patterns between different examples of the same category can be a very difficult task. However, the best models presented perform relatively well in the process of classifying the social relationship of the people in a couple into one of the four categories studied. In order to increase the effectiveness of the model, it is necessary to increase the number of examples in the database. In particular, it is necessary to increase the number of examples whose social relationship is that of a couple, %(see Table~\ref{tabla:Distribución_ejemplos}), 
as this is, in general, the relationship that shows the worst accuracy in the classification carried out by the models designed.

\subsection{Classification into two categories}
\label{Classification_in_two_categories}

Finally, in order to better analyse the results obtained, it was decided to merge the categories of couple, family and friendship into a single category called ``intimate``, in reference to the consideration that the three merged relationships entail a higher degree of intimacy between the members of the couple than the relationship of colleagues. In this sense, the category of colleagues is renamed as ``acquaintances`` for the purpose of the classification process.

In the following, we report the classification results obtained by the two types of NNs used. The characteristics of the different NN models implemented are detailed in Table~\ref{tabla:ResumenTF2-Binaria-Parámetros}. The results obtained by executing the networks defined in the Table~\ref{tabla:ResumenTF2-Binaria-Parámetros} are listed in Table~\ref{tabla:ResumenTF2-Binaria-Resultados}. To try to avoid the phenomenon of overfitting, L2 regularisation is implemented in some models. The confusion matrices of the different designed methods are collected in  table (\ref{tabla:RN2-7-2_tabla:RN2-8-2_tabla:RNR2-6-2}).
%tabla:RN2-7-2},~\ref{tabla:RN2-8-2},~\ref{tabla:RNR2-6-2}).

%\begin{table}[t]
%\begin{center}
%\begin{tabular}{| c | c | c | c | c | c | c | c | c |}
%\hline
%\parbox{0,25cm}{\textbf{}} & \parbox{0,83cm}{\textbf{Nº of hidden layers}} & \parbox{1,27cm}{\textbf{Nº of \\ neurons in \\ the hidden layers}} & \parbox{0,6cm}{\textbf{Nº of epochs}} & \parbox{0,90cm}{\textbf{Learning \\ rate}} & \parbox{0,20cm}{\textbf{L2 \\ Reg.}} & \parbox{0,8cm}{\textbf{Dropout}}\\
%\hline
%% \textbf{RN2-6} & 2 & 25-12 & 700 & 0,00015 & No & No \\ \hline
%\textbf{RN2-6} & 2 & 1500-600 & 2500 & 0,00011 & No & No \\ \hline
%\textbf{RN2-7} & 2 & 1500-600 & 2500 & 0,00011 & Yes & No \\ \hline
%%\textbf{RN2-8} & 2 & 1500-600 & 2500 & 0,00011 & Yes & Yes \\ \hline
%\textbf{RNR2-4} & 2 & 25-12 & 500 & 0,00011 & Yes & No \\ \hline
%\end{tabular}
%\caption{\small{Features of NN models implemented for the classification into two categories.}}
%\vspace{-6mm}
%\label{tabla:ResumenTF2-Binaria-Parámetros}
%\end{center}
%\end{table}

\begin{table}[t]
\begin{center}
\begin{tabular}{| c | c | c | c | c | c | c | c |}
\hline
\parbox{0,25cm}{\textbf{}} & \parbox{0,83cm}{\textbf{Nº of hidden layers}} & \parbox{1,27cm}{\textbf{Nº of \\ neurons in \\ the hidden layers}} & \parbox{0,8cm}{\textbf{Nº of epochs}} & \parbox{0,95cm}{\textbf{Learning \\ rate}} & \parbox{0,4cm}{\textbf{L2 \\ Reg.}}\\
\hline
% \textbf{RN2-6} & 2 & 25-12 & 700 & 0,00015 & No & No \\ \hline
\textbf{RN2-6} & 2 & 1500-600 & 2500 & 0,00011 & No \\ \hline
\textbf{RN2-7} & 2 & 1500-600 & 2500 & 0,00011 & Yes\\ \hline
%\textbf{RN2-8} & 2 & 1500-600 & 2500 & 0,00011 & Yes & Yes \\ \hline
\textbf{RNR2-4} & 2 & 25-12 & 500 & 0,00011 & Yes \\ \hline
\end{tabular}
\caption{\small{Features of NN models implemented for the classification into two categories.}}
\vspace{-6mm}
\label{tabla:ResumenTF2-Binaria-Parámetros}
\end{center}
\end{table}

\begin{table}[t]
\begin{center}
\begin{tabular}{ | c | c | c | c | c | c | c | c | c | }
\hline
\textbf{} & {\textbf{Training set accuracy}} & {\textbf{Test set accuracy}}\\
\hline
% \textbf{RN2-6} & 67,05\% & 66,67\% \\ \hline
\textbf{RN2-6} & 96,15\% & 71,26\% \\ \hline
\textbf{RN2-7} & 95,38\% & 63,22\% \\ \hline
%\textbf{RN2-8} & 75,38\% & 70,11\% \\ \hline
\textbf{RNR2-4} & 75,76\% & 65,38\% \\ \hline
\end{tabular}
\caption{\small{Accuracy of NN models implemented for the classification into two categories.}}
\vspace{-6mm}
\label{tabla:ResumenTF2-Binaria-Resultados}
\end{center}
\end{table}

\begin{table*}[t]
\begin{center}
\begin{tabular}{ | c | c | c | c | | c | c || c | c | }
\cline{3-8}
\multicolumn{2}{c|}{} &\multicolumn{6}{c|}{Predicted value} \\
\cline{3-8}
\multicolumn{2}{c|}{} &\multicolumn{2}{c||}{Model RN2-6}  &\multicolumn{2}{c||}{Model RN2-7} &\multicolumn{2}{c|}{Model RNR2-4}\\
\cline{3-8}
\multicolumn{2}{c|}{} & \textbf{Acquaintances} & \textbf{Intimate} & \textbf{Acquaintances} & \textbf{Intimate} & \textbf{Acquaintances} & \textbf{Intimate}\\
\hline
\multirow{2}{0,5cm}{Real value} & \textbf{Acquaintances} & \cellcolor{orange!44}45,83 & \cellcolor{orange!51}\textbf{54,17} & \cellcolor{orange!34}33,33 & \cellcolor{orange!57}\textbf{66,67}  & \cellcolor{orange!13}17,39 & \cellcolor{orange!70}\textbf{82,61}\\ 

\cline{2-8}
& \textbf{Intimate} & \cellcolor{orange!13}19,05 & \cellcolor{orange!70}\textbf{80,95} & \cellcolor{orange!23}25,40 & \cellcolor{orange!63}\textbf{74,60} & \cellcolor{orange!13}14,55 & \cellcolor{orange!70}\textbf{85,45} \\ \cline{1-8}
\end{tabular}
\caption{\small{Confusion matrix (in \%) of the RN2-6 model at left, RN2-7 model at center, and  RNR2-4 model at right.}}
\vspace{-6mm}
\label{tabla:RN2-7-2_tabla:RN2-8-2_tabla:RNR2-6-2}
\end{center}
\end{table*}

Table~\ref{tabla:RN2-7-2_tabla:RN2-8-2_tabla:RNR2-6-2}-{\it left} highlights that, using the RN2-6 model, the examples in the categories of acquaintances and intimate are correctly classified with accuracies of $45,83$ \% and $80,95$ \%, respectively. Table~\ref{tabla:RN2-7-2_tabla:RN2-8-2_tabla:RNR2-6-2}-{\it center} shows that the examples in the intimate category are correctly classified with an accuracy of $74,60$ \% when using the RN2-7 model. Table~\ref{tabla:RN2-7-2_tabla:RN2-8-2_tabla:RNR2-6-2}-{\it right} highlights that, using the RNR2-4 model, examples in the intimate category are correctly classified with an accuracy of $85,45$\%.

The classification into two possible categories shows a considerable increase in the accuracy in the test set of the different models designed, as can be seen in Table~\ref{tabla:ResumenTF2-Binaria-Resultados}. Specifically, the RN2-6 model stands out, with an accuracy of $96,15$ \% in the training set and $71,26$ \% in the test set. Furthermore, it can be seen in Table~\ref{tabla:RN2-7-2_tabla:RN2-8-2_tabla:RNR2-6-2}-{\it left} that the model has a high accuracy in the recognition of examples in the intimate category and by far the best accuracy of all models for the acquaintances category. In contrast, other models such as RN2-7 or RNR2-4 show a very considerable bias towards the intimate category, as can be seen in Tables~\ref{tabla:RN2-7-2_tabla:RN2-8-2_tabla:RNR2-6-2}-{\it center} and~\ref{tabla:RN2-7-2_tabla:RN2-8-2_tabla:RNR2-6-2}-{\it right}. Therefore, the RN2-6 model is postulated as the NN-based model capable of performing a better classification of human relationships into the two categories presented.

Thus, it can be stated that the process of classifying the social relationship into two categories is more accurate than the classification into four categories. In addition, as discussed above, it is necessary to increase the database of examples and to continue the process of setting hyperparameters in order to achieve a model with the best possible efficiency.

%This is due to the merging of the three categories that may be more difficult to distinguish into one category. However, the difficulty in differentiating between the categories of couple, family and friendship and the category of colleagues remains a problem in the classification process.

%\vspace{-4mm}
\subsection{Comparison with state-of-art-method}
\label{Comparison}

%\TODO{remarcar que obtenemos mejores resultados que la clasificacion anterior.}

Once the results of the different classification models have been obtained, they are compared with the results obtained in the work~\cite{yucel2019identification}. However, it is important to bear in mind that, although the objective of this project coincides with that of the comparative study, the databases used are not exactly the same. Nevertheless, it is possible to make a comparison of the results obtained by using different classification methods.

As various methods, parameters and even data are used in the study~\cite{yucel2019identification}, the best results obtained by some of the methods presented are used to compare them with the results obtained by the models developed in this work. Specifically, we select one of the methods of the study that obtains the best classification accuracies in the four categories of relationships (see Table~\ref{tabla:Resultados_Zanlungo}).

\begin{table}[t]
\begin{center}
\begin{tabular}{ | c | c | c | c | c | c | }
\cline{3-6}
\multicolumn{2}{c|}{} &\multicolumn{4}{c|}{Predicted value} \\
\cline{3-6}
\multicolumn{2}{c|}{} & \textbf{Colleagues} & \textbf{Couple} & \textbf{Family} & \textbf{Friendship}\\
\hline
\multirow{4}{0,5cm}{Real value} & \textbf{Colleagues} & \cellcolor{orange!57}\textbf{68,31} & \cellcolor{orange!5}7,29 & \cellcolor{orange!5}5,37 & \cellcolor{orange!13}19,03 \\ \cline{2-6}
& \textbf{Couple} & \cellcolor{orange!13}18,10 & \cellcolor{orange!34}\textbf{38,92} & \cellcolor{orange!23}20,66 & \cellcolor{orange!23}22,32 \\ \cline{2-6}
& \textbf{Family} & \cellcolor{orange!13}13,58 & \cellcolor{orange!34}31,11 & \cellcolor{orange!34}\textbf{36,57} & \cellcolor{orange!13}18,75 \\ \cline{2-6}
& \textbf{Friendship} & \cellcolor{orange!34}34,19 & \cellcolor{orange!13}16,66 & \cellcolor{orange!13}12,74 & \cellcolor{orange!34}\textbf{36,41} \\ \cline{1-6}
\end{tabular}
\caption{\small{Confusion matrix of one of the implemented methods in~\cite{yucel2019identification} (in \%).}}
\vspace{-4mm}
\label{tabla:Resultados_Zanlungo}
\vspace{-4mm}
\end{center}
\end{table}

Therefore, we compare the results of one of the best models of the study~\cite{yucel2019identification}, shown in Table~\ref{tabla:Resultados_Zanlungo}, with the results of the RN2-3 model developed in this work (see Table~\ref{tabla:RN2-3-2}), which is considered the model designed to best classify the social relationship in human couples. Thus, it is observed that while the method of the study presents a better accuracy in the categories of colleagues and couples, the developed model has a higher accuracy in the categories of family and friendship. However, the method implemented in~\cite{yucel2019identification} presents greater differences in the classification accuracies of the categories of colleagues and couple with respect to the method implemented in this work, than those presented by this method in the categories of family and friendship with respect to those of the comparative study. Consequently, it is difficult to say with certainty which model performs a better overall classification of social relationships in the four categories presented.

Even so, the potential offered by the deep learning models suggests that an increase in the database of examples of accompanying couples and some minor adjustments in the configuration of the hyperparameters of the model designed could provide the necessary improvement to obtain greater efficiency in the classification process. 

Finally, our new strategy to merge the three most similar categories in one, derived to obtain better classifications results than in the state-of-art method. And for a robot application to customize the robot's behavior to accompany people, this classification in two categories can be enough and derive in good accompaniment customization results.

\section{Discussion}
\label{Discussion} 

\subsection{Summary}
\label{Summary} 

We have designed a deep learning model that allows to classify the social relationship between two people  who are doing an accompaniment process into four possible categories (colleagues, couple, family or friendship) with a  good accuracy. The model has been developed using NNs and it has been trained and evaluated using a database of readings obtained from couples performing an accompaniment process in an urban environment. The confusion matrix has been used to verify the effectiveness of the classification and the results have been compared with those obtained in a similar work~\cite{yucel2019identification}.

We believe that the classification model can be improved to provide its full potential by increasing the database and adjusting the value of the hyperparameters, which can lead to its further adaptation to be implemented in a social robot.

\subsection{Contributions for HRI}
\label{Contributions} 

One of the pillars that technology must always bear in mind is to make people's lives easier. Thus, interaction between robots and people must be natural and safe. To achieve this, a robot must be able to identify how the person, it is accompanying  and interacting, is behaving.

By correctly identifying the behavior of the person it is accompanying, the social robot should be able to adapt its movement to generate a greater sense of security and comfort for the person and, therefore, improve their experience during the process.

\subsection{Applicability}
\label{Applicability} 

If the database is enhanced and some adjustments in the model are done in order to improve the efficiency, the described model  could be adapted to make the robot one of the members of the couple. Then, this new model could be implemented in a social robot. If this hypothetical social robot were to carry out a accompaniment process with a person, it should be able to act as an active member of that couple, navigating alongside them and classifying, in real time, the behavior of the accompanying human based on the measurements of its sensors  to adapt its navigation. % to the observed relationship that the companion presents with it.
In addition, the accompanied person could choose, prior to the accompaniment process, what type of relationship he/she wants the robot to have with him/her.%, for example, as if the two had a close relationship or, on the contrary, to act as would be expected of a work colleague.

It is important to take into account the advantage that a robotic system can adapt itself as comfortably as possible to a person while accompanying them to a destination. This can be very beneficial, for example, when using robot assistants for people with special needs,  sick or  elderly in nursing homes. Such robots can also support care home workers, helping to reduce their heavy workload. Ultimately, providing robots with the tools that allow them to adapt their behavior in a personalised way, and behave in the most social way possible can only bring benefits for the society of the future.

\subsection{Limitations}
\label{Limitations} 

Although social relationship classification in couples may prove to be a useful application, we believe that the maximum potential and most applications will come from adapting this function so that a social robot can benefit from and be able to adapt its behavior to the preferences shown by the person it is accompanying.

\section{Conclusions}
\label{sec:conclusions} 
We have presented several models capable of  classifying the social behavior of a couple of humans during their accompaniment. From the analysis of the results, the RN2-3 model based on a standard NN stands out, with which good accuracies are obtained in the process of classifying the social relationship between two people.

The future work would be oriented towards two different objectives. The first objective would be trying to improve the precision of the implemented NNs. To achieve this, it is essential to increase the database with more labeled examples of couples accompanying each other and, then, continue adjusting the parameters of the models and modifying their design to increase their effectiveness.

The second objective to be carried out would be the implantation of the designed models in a real robot with the necessary instruments to collect data from the target couples. If this phase were to be carried out successfully, the next step to be taken would be to adapt the programs designed to consider the robot as one of the members of the couple. Then, the robot should obtain the data of various parameters of the human companion's navigation while it is moving to be able to process them through the new models implemented and, thus, classify its behavior in one of the four categories studied. Finally, the robot would have the necessary information to be able to adapt to the behavior shown by its human companion. Using this process, the robot should be able to accompany people in a more natural, safe, social and comfortable way for them.

%\section*{References}

%\newpage
\bibliographystyle{IEEEtran}
\balance 
\bibliography{IEEEabrv,tfmBib}
%\begin{thebibliography}{00}

%\end{thebibliography}
\vspace{12pt}

\end{document}